\documentclass[11pt]{article}

\usepackage[margin=1in]{geometry}
\usepackage{graphicx}
\usepackage{booktabs}
\usepackage{array}
\usepackage{url}
\usepackage{xcolor}
\usepackage{hyperref}
\usepackage{amsmath}

\hypersetup{
  colorlinks=true,
  linkcolor=blue,
  citecolor=blue,
  urlcolor=blue
}

\title{Do AI-Native Biotechs Need Departments?\\
\large Benchmarking Company World Models for AI-Driven Drug Development}

\author{Yinan Wang\\
Noah AI Research}
\date{\today}

\begin{document}
\maketitle

\begin{abstract}
AI-native biotechnology companies are often designed by copying the human
biotech org chart into agent roles: biology, clinical development, regulatory
affairs, manufacturing, business development, commercial, project management,
and investment committee. We argue that this is incomplete as the core
computational abstraction.
Recent work on world models frames capable agents as systems that maintain
predictive state representations, simulate action-conditioned futures, and
revise those models when observations disagree with expectations. Translating
that idea to biopharma, we define a \emph{Company World Model}: a persistent,
auditable model of asset state, evidence, uncertainty, regulatory path,
partner/buyer demand, commercial potential, resources, and the expected effect
of candidate actions on enterprise value. We introduce a dry-lab benchmark for
testing whether AI-agent organizations should mimic departments or operate
around such an asset-to-value world model. The benchmark contains 45
retrospective public-information decision cases with strict time cutoffs,
hidden outcomes, common output schemas, automatic scoring, and blinded
pairwise judging. We compare human-org-mimic, stronger human-org-mimic-plus,
AI-native asset-centric, and AI-native value-conversion architectures. The
value-conversion architecture is a prompt-level approximation of a Company
World Model: a Live Asset Value Record is updated by Deal, Approval, Revenue,
and Investment Arbiter loops. Under a success function defined by external
business development (BD), regulatory approval and launch, and revenue
discipline, this architecture achieved the highest automatic score among the
original three arms (4.68 vs. 4.33 for human-org mimic and 4.26 for
asset-centric AI-native) and was preferred strongly over the original
baselines by value-specific blinded judges. Stress tests narrowed the claim: a
stronger human baseline improved to 4.49, value-conversion AI-native beat it
only directionally in Codex value-specific judging (26--19), and a neutral
decision-quality judge slightly preferred human-org-mimic-plus (25--20).
Mechanistic ablations are consistent with, but do not prove, the world-model
interpretation: the full architecture was preferred over versions without Revenue Room
(44--1), Deal Room (41--4), Approval Room (35--10), and Investment Arbiter
(27--18). The central finding is objective-sensitive: an AI-native biopharma
should expose the objective function its world model optimizes. Departments may
remain useful governance views, but the core operating primitive should be a
shared, predictive asset-to-value state rather than a static human org chart.
The study is dry-lab only and does not establish real-world drug success or
revenue prediction accuracy.
\end{abstract}

\section{Introduction}

Biopharma is not only a scientific discovery problem. It is a sequential
decision problem under uncertainty. A company must decide which targets to
pursue, which assets to fund, what evidence would change belief, when to stop
development, when to partner, what evidence is sufficient for regulators, and
whether a product can become commercially meaningful after launch. Human
biotechnology and pharmaceutical companies usually solve this by separating
work into functions: biology, translational science, clinical development,
regulatory affairs, chemistry/manufacturing/controls (CMC), commercial, BD,
finance, project management, and investment-committee governance.

That structure exists for good reasons. Humans have limited context windows,
limited working memory, limited bandwidth, and specialized training.
Departments distribute cognition and accountability across people. AI-native
organizations have different constraints. AI agents can operate from a shared
state object, retrieve long histories, call tools dynamically, spawn
specialized subroutines, and synthesize scientific, regulatory, commercial,
and financial evidence into the same decision state. If those capabilities are
real, copying a human org chart may be a poor default design principle.

Recent world-model research offers a better language for this problem.
World-model agents maintain internal representations of environment state,
predict how state changes under actions, plan over possible futures, and
update the model when new observations contradict prior expectations
\cite{ha2018world,bruce2024genie,assran2025vjepa2,chu2026agenticworld}.
In physical AI, the state might be a scene and the actions might be robot
movements. In an AI-native biopharma, the state is an asset and company value
context: biology claims, evidence strength, uncertainty, IP position,
competitive landscape, regulatory path, partner demand, patient and payer
logic, launch assumptions, capital constraints, and execution options. The
actions are experiments, trial designs, regulatory interactions, BD outreach,
indication choices, evidence-package changes, financing choices, and kill or
partner decisions.

We call this abstraction a \emph{Company World Model}. It is not a dashboard
and not merely a knowledge graph. It is a persistent, auditable state-and-
transition model that asks: given the current asset state, what action is most
likely to improve the probability of external BD, regulatory approval and
launch, and expected revenue, under time, capital, evidence, and competitive
constraints? In this framing, the central organizational primitive is no
longer the department. It is the live asset-to-value state over which agents
simulate, challenge, execute, and update.

This paper asks a direct question: do AI-native biotechs need departments, or
should they be organized around a Company World Model? We study this question
in the dry lab. We do not run experiments, design new molecules, or claim that
an AI system can replace real clinical development. Instead, we construct a
retrospective decision benchmark. Each case asks an AI-agent organization to
make a historically grounded biopharma decision using only public sources
available before a specified cutoff date. The model does not see the later
outcome. The same case prompt, source set, output schema, and evaluation
framework are used across architectures. This allows us to vary one design
variable: how agents are organized.

The initial hypothesis was that an AI-native asset-centric architecture should
outperform a human-org-mimic architecture because it organizes work around a
single live asset record rather than a chain of department-like memos. Early
experiments partially supported this: the asset-centric architecture improved
risk recognition and actionability in failure and ambiguous clinical cases.
But it exposed a gap. A drug program does not succeed by having a
well-reasoned scientific record alone. In practice, a successful asset must
convert into external BD value, regulatory approval and launch, and expected
sales.

\begin{figure*}[t]
  \centering
  \includegraphics[width=0.95\linewidth]{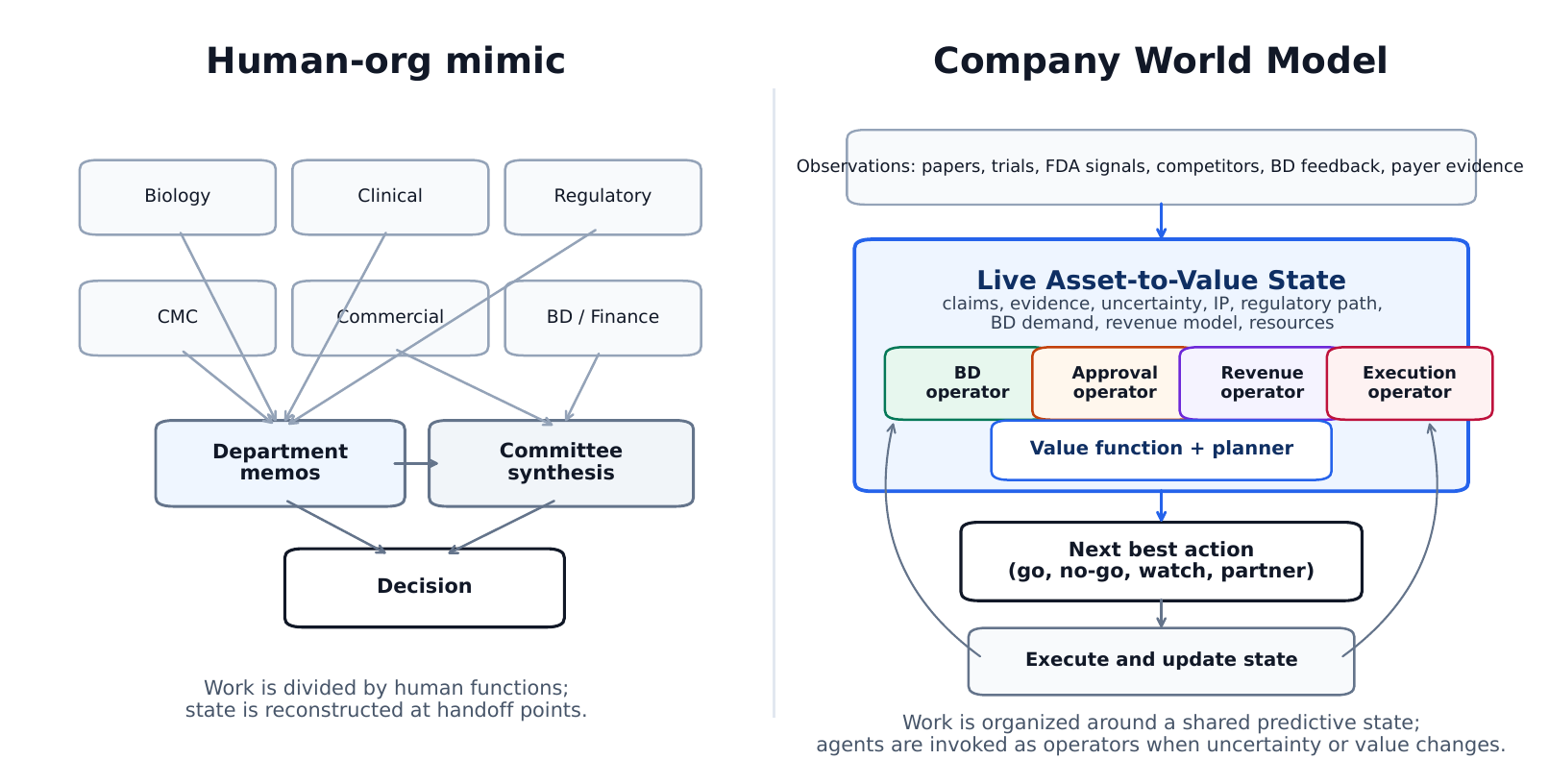}
  \caption{The benchmark compares a human-org-mimic architecture, in which
  functional agents produce department-like memos before committee synthesis,
  with a Company World Model architecture, in which agents operate on one
  shared asset-to-value state, simulate objective-specific transitions, and
  choose the next action through a planning layer.}
  \label{fig:architectures}
\end{figure*}

We therefore tested a third architecture: an AI-native value-conversion
organization. This design is our first prompt-level approximation of a Company
World Model. A Live Asset Value Record represents the current state. Deal,
Approval, and Revenue loops model how actions might change partner interest,
regulatory feasibility, label quality, payer access, adoption, and forecast
quality. An Investment Arbiter acts as a planner over the shared state,
identifying the binding constraint and recommending \texttt{go},
\texttt{no-go}, \texttt{watch}, or \texttt{partner}
(Figure~\ref{fig:architectures}).

The first value-conversion result was strong, but a skeptical reviewer would
ask whether this is circular: did we design a value-conversion architecture and
then evaluate it with a value-conversion rubric? We therefore added a stronger
human baseline, a neutral judge prompt, and mechanistic ablations. These tests
change the interpretation. The best current claim is not that
value-conversion AI-native is universally better. The best claim is that an
AI-native organization must expose the objective function its world model is
optimizing, and that department copying is not the only plausible baseline.

This paper contributes:
\begin{enumerate}
  \item a benchmark framing for evaluating AI-agent organizational design in a
  high-stakes scientific business domain;
  \item the Company World Model abstraction for AI-native biopharma: a shared
  asset-to-value state, transition model, objective function, and planner;
  \item a 45-case retrospective biopharma decision dataset with time cutoffs,
  hidden outcomes, common schemas, and source constraints;
  \item matched agentic organizational architectures: human-org mimic,
  human-org-mimic plus, AI-native asset-centric, and AI-native
  value-conversion/Company-World-Model;
  \item automatic and blinded pairwise evaluation showing that
  value-conversion organization helps under a BD/approval/revenue objective
  but does not robustly dominate under neutral judging;
  \item mechanistic ablations testing which value-conversion loops drive the
  full architecture's advantage under the target objective.
\end{enumerate}

\section{Related Work}

\paragraph{LLM agents and agentic architectures.}
Recent work on language-model agents has shown that models can be embedded in
systems with memory, planning, tool use, reflection, and iterative execution.
Generative Agents introduced an architecture for simulated agents with memory
retrieval, reflection, planning, and social interaction~\cite{park2023generative}.
Voyager demonstrated an open-ended embodied agent that accumulates reusable
skills and uses environment feedback~\cite{wang2023voyager}. ReAct,
Reflexion, Tree of Thoughts, HuggingGPT, AgentBench, AgentBoard, and
SWE-bench further illustrate that agent performance depends on decomposition,
tool use, execution environment, and evaluation design~\cite{yao2023react,
shinn2023reflexion,yao2023tree,shen2023hugginggpt,liu2023agentbench,
ma2024agentboard,jimenez2024swebench}. Our study applies this systems question
to organizational design.

\paragraph{World models for agentic planning.}
World models originated as learned internal models that allow agents to plan
or learn policies inside compressed representations of environment dynamics
\cite{ha2018world}. Recent work has expanded the idea from reinforcement
learning environments to foundation world models and action-controllable
interactive environments \cite{bruce2024genie}. V-JEPA 2 frames world models
as self-supervised systems that support understanding, prediction, and planning
in the physical world \cite{assran2025vjepa2}. Agentic World Modeling further
proposes a levels-by-laws taxonomy in which world models can be predictors,
simulators, or self-updating evolvers operating across physical, digital,
social, and scientific regimes \cite{chu2026agenticworld}. Our use of
``Company World Model'' is a domain translation of this state-transition-
planning view. A biopharma company does not need to predict pixels; it needs
to predict how actions change evidence, regulatory feasibility, BD demand,
commercial potential, execution risk, and value under uncertainty.

\paragraph{AI for scientific and chemical discovery.}
Autonomous scientific systems such as Coscientist have shown that LLM-based
agents can plan and execute chemical research workflows when connected to tools
and experimental infrastructure~\cite{boiko2023autonomous}. AlphaFold and
other scientific AI systems demonstrate the broader value of machine learning
for biomedical discovery~\cite{jumper2021alphafold}. The AI Scientist explores
automated scientific ideation, experiment execution, paper writing, and
reviewing~\cite{lu2024scientist}. These works focus primarily on hypothesis
generation, experiments, protocols, or scientific outputs. Our benchmark
focuses on dry-lab company and asset decisions.

\paragraph{Benchmarks and LLM-as-judge evaluation.}
Large-scale benchmark work such as BIG-bench and HELM has emphasized
standardized, transparent evaluation across tasks and models~\cite{
srivastava2022beyond,liang2022helm}. LLM-as-judge methods such as MT-Bench and
G-Eval provide scalable preference evaluation, but they introduce risks of
prompt, style, and model bias~\cite{zheng2023judging,liu2023geval}. Biomedical
benchmarks such as PubMedQA, MedQA, and Med-PaLM evaluate question answering or
clinical reasoning~\cite{jin2019pubmedqa,jin2021disease,singhal2023large}.
Biopharma asset decisions are different: outcomes unfold over years, and a
decision that later looks wrong may have been reasonable at the historical
cutoff. We therefore use time cutoffs, allowed-source lists, hidden outcomes,
acceptable-decision sets, automatic scoring, and blinded pairwise judging.

\paragraph{Organizational design for AI agents.}
Many agent systems implicitly copy human organizations by assigning agents to
roles such as scientist, engineer, manager, reviewer, or planner. In
biopharma, the mapping is especially tempting because human companies already
have well-defined functions. However, human organizations are designed around
human coordination costs. If agents can share state and dynamically load
skills, then the better organizational primitive may be an object, workflow, or
objective function rather than a department.

\section{Benchmark Design}

\subsection{Task Definition}

Each benchmark case presents a historically grounded biopharma decision. The
agent organization receives a case title and prompt, an as-of date, a list of
allowed public sources published on or before the cutoff, allowed decisions
(\texttt{go}, \texttt{no-go}, \texttt{watch}, or \texttt{partner}), and a common
JSON output schema. The model does not receive the benchmark category, hidden
outcome, gold decision, acceptable-decision set, or scoring notes. It must
return a structured decision with confidence, claims, evidence,
counterevidence, uncertainties, red flags, next actions, and an audit trail.

\subsection{Dataset}

The current benchmark contains 45 retrospective cases: 13 BD/competitive
cases, 10 failure cases, 12 mixed regulatory or evidence cases, and 10 success
cases. Examples include semaglutide in obesity, sotatercept in pulmonary
arterial hypertension, exa-cel in sickle cell disease, KarXT in schizophrenia,
selonsertib and obeticholic acid in NASH, bempegaldesleukin in melanoma,
aducanumab in Alzheimer's disease, Elevidys in Duchenne muscular dystrophy,
roxadustat in CKD anemia, Merck-Acceleron, Pfizer-Seagen, Gilead-Forty Seven,
AbbVie-ImmunoGen, and BMS-Karuna. All cases are built from public sources and
date-checked against the case cutoff. The public ancillary file included with
this submission removes hidden outcomes and scoring notes so the benchmark can
remain usable for future evaluation.

\subsection{Success Function and Company World Model}

For the value-conversion study, we define drug-program success as conversion
into one or more of three external outcomes:
\begin{enumerate}
  \item successful external BD, licensing, acquisition, or strategically
  valuable partnering;
  \item regulatory approval and launch;
  \item expected sales revenue.
\end{enumerate}
This is a deliberately narrow enterprise value-conversion objective, not a
complete normative definition of a drug's medical or social value. Safety,
clinical benefit, unmet need, standard of care, manufacturing feasibility, and
post-launch evidence remain constraints and state variables inside the model.
The third outcome is not evaluated as true forecast accuracy in this version.
The current benchmark measures revenue discipline as a proxy: whether the
output reasons about target patients, payer/access, label-driven market size,
physician adoption, competition, launch assumptions, forecast gates, and
sales-risk triggers.

We formalize the Company World Model at the level required for this dry-lab
study. Let $S_t$ denote the live company state for an asset at time $t$:
scientific claims, evidence, uncertainty, counterevidence, red flags,
regulatory path, IP and CMC constraints, partner/buyer demand, commercial
adoption assumptions, resources, and timing. Let $a_t$ denote a candidate
action such as generating evidence, changing indication, preparing a BD
package, pursuing a partner, seeking regulatory advice, delaying investment,
or stopping the program. A Company World Model contains:
\begin{enumerate}
  \item a state representation $S_t$;
  \item a transition model $\hat{T}(S_{t+1}\mid S_t,a_t)$ over how actions are
  expected to change evidence, uncertainty, regulatory feasibility, BD demand,
  commercial potential, cost, and time;
  \item a value function $V(S_t)$ tied to external BD, approval/launch, and
  revenue discipline;
  \item a planner $\pi(a_t\mid S_t)$ that selects the next action under
  uncertainty and resource constraints;
  \item an update process that revises $S_t$, $\hat{T}$, and $V$ when new
  evidence or market feedback arrives.
\end{enumerate}

The present benchmark does not train or validate a numerical simulator for
$\hat{T}$. Instead, it tests whether a prompt-level architecture that forces
the agent harness to reason in this state-transition-value-planning structure
produces better cutoff-aware decisions than department-mimicking baselines.

\subsection{Architectures}

\paragraph{Human-org mimic.}
The human-org-mimic architecture mirrors a conventional biotechnology company.
Functional agents reason as biology, clinical, regulatory, CMC, commercial,
BD/finance, project-management, and investment-committee roles. The final
decision is synthesized through a project-manager or investment-committee
style layer.

\paragraph{Human-org-mimic plus.}
The human-org-mimic-plus architecture is a stronger human baseline added after
reviewer-style critique. It keeps human departments but gives them a shared
asset memory, explicit BD/commercial challenge, and investment-committee
conflict resolution. It is designed to avoid a weak strawman comparison.

\paragraph{AI-native asset-centric.}
The AI-native asset-centric architecture organizes work around a live asset
record. Instead of passing department memos, agents update one shared object
containing claims, evidence, counterevidence, uncertainties, red flags, and
next-best actions.

\paragraph{AI-native value conversion / Company World Model.}
The AI-native value-conversion architecture extends the asset-centric approach
by aligning the organization with the three external success outcomes. In
world-model terms, the Live Asset Value Record is the state representation.
Deal Room, Approval Room, and Revenue Room act as domain transition/value
models for partner demand, regulatory/label feasibility, and commercial
conversion. The Investment Arbiter is the planner that identifies the binding
constraint and chooses the next capital-allocation action. These rooms are not
intended to be new departments. They are views and operators over one shared
asset-to-value state.

\subsection{Execution Harness}

The empirical outputs were generated with Codex CLI as the agent harness using
model \texttt{gpt-5.5}. Runs used hidden-outcome-free prompt packs and a
read-only sandbox. The comparison should therefore be interpreted as an
agent-harness benchmark, not as a raw API model benchmark. All architectures
used the same case files, allowed sources, output schema, and downstream
scoring pipeline. The three-arm run produced 135 outputs. The
human-org-mimic-plus stress test added 45 outputs, producing a four-arm
comparison with 180 outputs. The value-conversion ablation run added four
single-component removal arms across the same 45 cases, producing 180
additional ablation outputs and a 225-output five-arm mechanism dataset. Three
ablation items timed out under the default local Codex configuration and were
rerun with \texttt{codex exec --ignore-user-config}; all three were still
generated by the same prompt-pack runner and constrained to the same output
schema.

\subsection{Evaluation}

We use four evaluation layers. First, automatic scoring computes decision
quality, evidence quality, actionability, efficiency, auditability,
cutoff-aware acceptable-decision match, exact retrospective decision match,
red-flag recall, and value-conversion proxy scores for BD, approval, and
revenue discipline. Second, value-specific blinded pairwise judges compare
outputs A and B without architecture labels and are asked which output better
converts the asset into BD, approval, and revenue value under
cutoff-appropriate uncertainty. Third, a neutral judge stress test uses a more
general prompt prioritizing decision quality, evidence and counterevidence,
risk recognition, actionability, and auditability. Fourth, a mechanistic
ablation compares the full value-conversion architecture against four versions
that remove exactly one loop: Deal Room, Approval Room, Revenue Room, or
Investment Arbiter.

\section{Results}

\subsection{The Original Asset-Centric AI-Native Architecture Was Incomplete}

Before testing the value-conversion architecture, we compared human-org mimic
against the original asset-centric AI-native architecture on the 45-case
benchmark under the BD/approval/revenue objective. Automatic value-conversion
scoring slightly favored human-org mimic (4.33 vs. 4.26). Blinded
value-conversion judging was closer and slightly favored asset-centric
AI-native overall: Codex preferred it 23--22, and Claude Opus 4.8 preferred it
26--19. However, both judges preferred human-org mimic on BD/competitive cases
by 8--5. This suggested that the first AI-native architecture was useful for
evidence and risk management, but incomplete as a company-level operating
model.

\subsection{Three-Arm Value-Conversion Result}

\begin{figure}[t]
  \centering
  \includegraphics[width=0.78\linewidth]{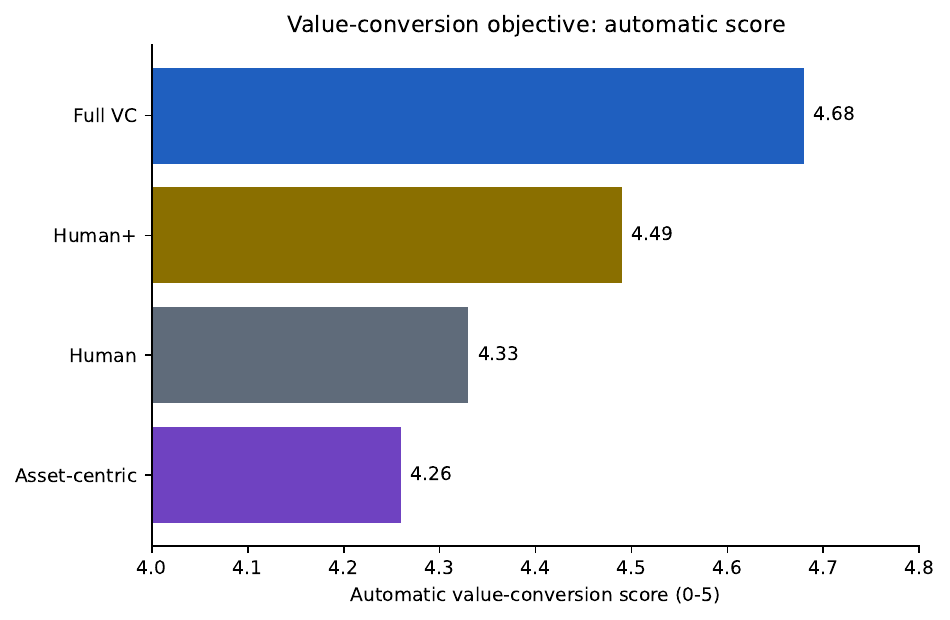}
  \caption{Automatic value-conversion score after adding the stronger
  human-org-mimic-plus baseline. The score is a proxy for BD, approval, and
  revenue-conversion discipline; blinded judging is the primary preference
  result when available.}
  \label{fig:value_scores}
\end{figure}

When the explicitly value-conversion-centric architecture was added as a third
arm, it achieved the highest standard benchmark score and the highest
value-conversion proxy score (Table~\ref{tab:threearm};
Figure~\ref{fig:value_scores}). Value-specific blinded judges preferred
value-conversion AI-native strongly over the original baselines: Codex
preferred it 42--3 over human-org mimic and 44--1 over asset-centric
AI-native; Claude preferred it 40--5 and 42--3 respectively. This is the
strongest positive evidence for value-conversion organization, but it is also
the result most vulnerable to rubric-architecture circularity.

\begin{table}[t]
\centering
\caption{Automatic value-conversion proxy scores in the original three-arm
comparison. Scores are on a 0--5 scale except decision credit.}
\label{tab:threearm}
\begin{tabular}{lccccc}
\toprule
Architecture & Value & BD & Approval & Revenue & Decision \\
\midrule
AI-native value conversion & 4.68 & 4.65 & 4.72 & 4.54 & 0.91 \\
Human-org mimic & 4.33 & 3.86 & 4.65 & 3.91 & 0.88 \\
AI-native asset-centric & 4.26 & 3.77 & 4.60 & 3.77 & 0.86 \\
\bottomrule
\end{tabular}
\end{table}

\subsection{Strong Human Baseline and Neutral Judge Stress Tests}

We added \texttt{human\_org\_mimic\_plus}, a stronger baseline with shared
asset memory, explicit BD/commercial challenge, and investment-committee
synthesis. Automatic value-conversion scoring improved substantially versus
the original human-org mimic (4.49 vs. 4.33), narrowing the gap with
value-conversion AI-native (4.68). Under value-specific Codex judging,
value-conversion AI-native beat human-org-mimic-plus 26--19. This is
directionally positive but not statistically strong under an exact two-sided
binomial test against a 50--50 null ($p=0.371$).

We then reran blinded judging with a neutral decision-quality prompt. The
neutral prompt did not foreground BD/approval/revenue value conversion. Under
this judge, value-conversion AI-native no longer dominated: human-org mimic
beat it 23--22, human-org-mimic-plus beat it 25--20, and asset-centric
AI-native beat it 23--22. The human-plus neutral result is also not
statistically strong ($p=0.551$ for 25/45). The important point is not that the
neutral judge proves human-like departments are better; it is that the
value-conversion advantage does not robustly transfer to a neutral
general-quality objective.

\subsection{Mechanistic Value-Conversion Ablations}

We next tested whether the full value-conversion architecture depends on its
named loops. Four ablations removed one component at a time while retaining
the Live Asset Value Record and remaining rooms. Automatic value-conversion
scoring ranked the full architecture highest (Table~\ref{tab:ablation_scores}).
Removing Deal Room produced the largest BD-conversion drop. Removing Revenue
Room produced the largest revenue-discipline drop. Removing Approval Room
lowered approval conversion and actionability. Removing Investment Arbiter had
the smallest automatic-score effect.

\begin{table}[t]
\centering
\caption{Automatic value-conversion scores for the full architecture and four
single-component removals.}
\label{tab:ablation_scores}
\begin{tabular}{lccccc}
\toprule
Architecture & Value & BD & Approval & Revenue & Decision \\
\midrule
Full value conversion & 4.68 & 4.65 & 4.72 & 4.54 & 0.91 \\
No Investment Arbiter & 4.59 & 4.42 & 4.64 & 4.52 & 0.87 \\
No Approval Room & 4.50 & 4.43 & 4.46 & 4.52 & 0.85 \\
No Deal Room & 4.47 & 3.77 & 4.68 & 4.47 & 0.88 \\
No Revenue Room & 4.32 & 4.49 & 4.59 & 3.55 & 0.86 \\
\bottomrule
\end{tabular}
\end{table}

\begin{figure}[t]
  \centering
  \includegraphics[width=0.92\linewidth]{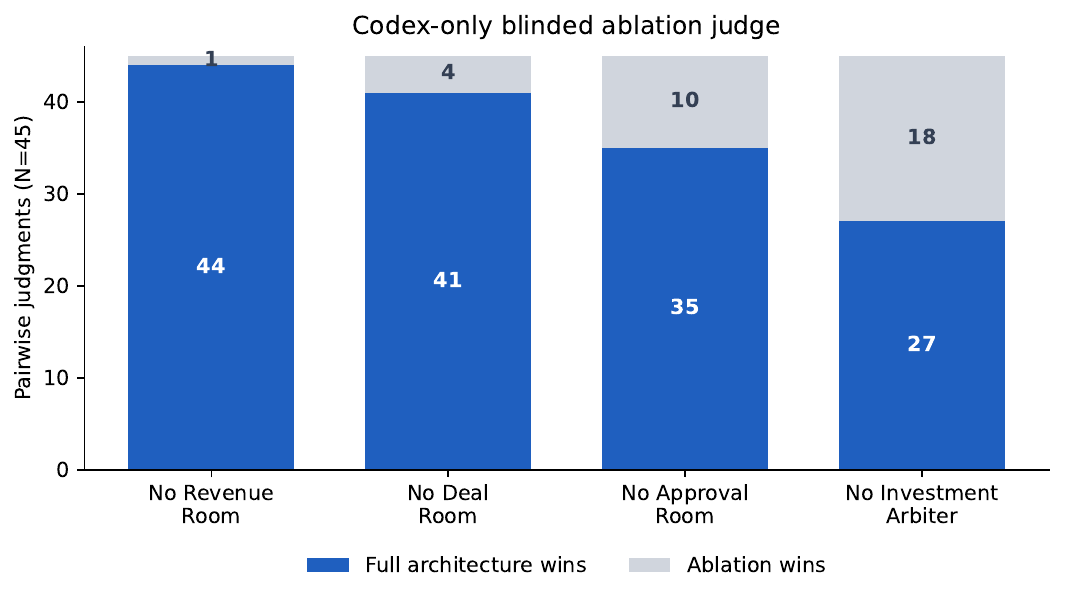}
  \caption{Codex-only blinded value-conversion judge preferences for the full
  architecture versus single-component ablations.}
  \label{fig:ablation}
\end{figure}

Blinded Codex value-conversion judging also preferred the full architecture in
all four ablation comparisons (Figure~\ref{fig:ablation}; Table~\ref{tab:prefs}).
The strongest signals are no Revenue Room (44--1, $p=2.61 \times 10^{-12}$),
no Deal Room (41--4, $p=9.33 \times 10^{-9}$), and no Approval Room (35--10,
$p=2.47 \times 10^{-4}$). The no-Investment-Arbiter comparison is weaker
(27--18, $p=0.233$) and should be interpreted as directional only.

\begin{table}[t]
\centering
\caption{Blinded preference statistics. Wilson intervals are 95\% intervals
for the proportion of full-architecture wins.}
\label{tab:prefs}
\begin{tabular}{lcccc}
\toprule
Comparison & Full wins & N & Win rate & Exact $p$ \\
\midrule
Full vs. no Revenue Room & 44 & 45 & 0.978 & $2.61{\times}10^{-12}$ \\
Full vs. no Deal Room & 41 & 45 & 0.911 & $9.33{\times}10^{-9}$ \\
Full vs. no Approval Room & 35 & 45 & 0.778 & $2.47{\times}10^{-4}$ \\
Full vs. no Investment Arbiter & 27 & 45 & 0.600 & 0.233 \\
Full vs. human-org-mimic-plus & 26 & 45 & 0.578 & 0.371 \\
\bottomrule
\end{tabular}
\end{table}

These ablations support a mechanistic reading: the full architecture does not
only win by using a different writing template. It loses measurable capability
when specific conversion loops are removed. However, this ablation judge is
currently Codex-only, so cross-model and human expert replication remain
necessary.

\subsection{Output Length and Case-Level Checks}

A common concern in pairwise LLM judging is that judges may prefer longer or
more structured outputs. Mean output lengths in the original three-arm run
were similar: 12,144 characters for AI-native asset-centric, 12,115 for
AI-native value conversion, and 11,821 for human-org mimic. This does not
eliminate all style or template bias, but a simple ``longer output wins''
explanation is insufficient. We also created qualitative case studies for
Pfizer-Seagen, Amgen-Horizon, and Merck-Prometheus to illustrate cases where
value-specific and neutral judges agree or disagree.

\subsection{Interpretation Stress Tests}

We ran two adversarial interpretation checks after reframing the architecture
as a Company World Model. The first check asks whether the paper overclaims
general superiority. The answer is no: the neutral judge results explicitly
reject a universal-dominance interpretation, and the strongest positive claim
is limited to the BD/approval/revenue objective. The second check asks whether
``world model'' is merely a relabeling of a prompt template. The current
evidence is intermediate. The ablations show that removing domain operators
damages the architecture in directionally specific ways, which is consistent
with the state-transition-value-planning interpretation. However, because the
system is implemented as prompt-level role structure rather than an
independently trained simulator, the benchmark should be read as testing a
world-model-inspired organizational architecture, not as validating a learned
biopharma world model.

\section{Discussion}

\subsection{The Company World Model Is Objective-Sensitive}

The central finding after stress testing is objective sensitivity. If the
evaluation target is BD, approval, and revenue conversion, organizing agents
around Deal, Approval, Revenue, and Investment Arbiter loops appears useful. If
the evaluation target is neutral decision quality, evidence, risk, and
auditability, the same architecture does not robustly beat strong department-
or asset-centric baselines. In world-model language, the learned lesson is
that the value function matters. A Company World Model optimized for
asset-to-value conversion will not necessarily maximize a neutral diligence
rubric, just as a physical world model trained for navigation is not
automatically optimal for manipulation. AI-native biopharma systems should
therefore expose the objective function they are optimizing rather than hide it
inside agent-role prompts.

\subsection{Rooms Are Operators, Not Departments}

The first draft risked making human departments look like a weak strawman. The
human-org-mimic-plus baseline shows why that would be wrong. Adding shared
asset memory, BD/commercial challenge, and investment-committee conflict
resolution substantially improved the human-like organization. The real
contrast is not ``departments vs. no departments.'' It is fixed functional
ownership versus objective-aligned operators over shared state.

This distinction is important for product and organizational design. A Deal
Room should not become a renamed BD department. It should be a transition
model over buyer demand, transaction feasibility, strategic fit, valuation
gates, and walk-away conditions. An Approval Room should not be a renamed
regulatory department. It should model how evidence, endpoint choice, safety,
CMC, and label assumptions change approval and launch feasibility. A Revenue
Room should not be a late-stage commercial memo. It should model patient
selection, access, adoption, competition, forecast assumptions, and launch
risk from the beginning of asset planning.

\subsection{Why Asset-Centric Alone Was Insufficient}

The asset-centric AI-native architecture improved traceability and risk
recognition because it forced claims, evidence, counterevidence,
uncertainties, and next actions into one live record. That is a necessary
state layer, but not a complete company world model. Without transition and
value layers, an asset record can become a better notebook rather than a
better operating system. The ablation results support this interpretation:
removing Revenue Room mostly damaged revenue discipline, removing Deal Room
damaged BD conversion, and removing Approval Room damaged approval conversion.
The strongest product implication is that commercial and deal conversion
cannot be appended after scientific diligence; they must be part of the state
model from the start.

\subsection{Implications for AI-Native Biopharma Products}

The benchmark suggests a product architecture with six layers:
\begin{enumerate}
  \item an observation layer that ingests papers, patents, clinical data,
  competitor events, regulatory signals, BD feedback, payer evidence, and
  internal execution updates;
  \item a Live Asset Value State containing evidence, assumptions, risks,
  uncertainty, BD options, approval path, commercial model, resources, and
  timeline;
  \item domain transition operators for biology, evidence generation,
  regulatory/label feasibility, deal demand, launch/revenue, and execution
  risk;
  \item a value function that makes the current optimization target explicit,
  for example BD readiness, approval probability, revenue discipline, capital
  efficiency, or neutral scientific diligence;
  \item a planner and Investment Arbiter that select next actions and allocate
  resources based on the binding constraint;
  \item an execution network that runs analyses, drafts BD packages, updates
  forecasts, designs evidence plans, and writes regulatory or commercial work
  products, while feeding observations back into the state.
\end{enumerate}
This architecture makes AI capability growth easier to absorb. As base models
gain longer context, better tool use, stronger memory, and more reliable
planning, the product can improve the transition operators and planner without
redrawing a static org chart. Departments can remain as human-facing
governance views, audit boundaries, or expert-review interfaces, but they are
not the deepest computational primitive.

\section{Limitations}

This study is dry-lab only. It does not show that an AI-native organization
can discover better drugs, run better trials, win better labels, negotiate
real deals, or produce higher sales. It evaluates retrospective decision
outputs under public-source constraints.

The Company World Model is implemented here as a prompt-level architecture,
not as a learned numerical simulator. We do not validate transition estimates
against realized future changes in asset value, regulatory probability, BD
terms, or sales. Revenue is measured as reasoning discipline, not forecast
accuracy. The benchmark tests whether outputs reason about payer access,
launch adoption, competition, and forecast assumptions; it does not compare
pre-cutoff forecasts with actual post-launch sales.

The outputs are generated with Codex CLI as an agent harness, so the results
should not be described as raw API model performance or as universal
properties of a base model. The new strong-baseline, neutral-judge, and
mechanistic-ablation stress tests currently use Codex judging only. A Claude
neutral replication was started but blocked by a \texttt{403 Request not
allowed} authentication/API error after 43 of 135 items. Human expert review
by BD, clinical development, regulatory, and commercial specialists remains
necessary.

Finally, the value-conversion architecture is more aligned with the
value-conversion judging prompt than the baselines. The neutral judge stress
test shows that this alignment matters. Future work should use multiple
orthogonal judging prompts, human reviewers, stricter output-length controls,
and longitudinal cases where the world model is updated over several simulated
decision points.

\section{Conclusion}

Do AI-native biotechs need departments? This benchmark suggests that the
answer is not a simple no, but it also should not be yes by default. The
original three-arm experiment supported the hypothesis that an AI-native
value-conversion architecture can outperform human-org mimic and asset-centric
AI-native architectures under a BD/approval/revenue objective. Stronger
baselines and neutral judging narrow the claim. Departments are not obsolete as
human governance interfaces, but they are incomplete if copied as the core
computational architecture.

The best current conclusion is that an AI-native biopharma should be designed
around a Company World Model: a shared, auditable asset-to-value state;
transition operators for science, regulation, BD, commercial, and execution
dynamics; an explicit value function; and a planner that selects the next
action under uncertainty. This result is less sweeping than the slogan
``AI-native beats departments,'' but more useful: it turns AI-native
organization from a metaphor into a testable systems-design variable.

\section*{Artifact Availability and Reproducibility}

The submission package includes a public no-label version of the 45 benchmark
cases, figure files, statistical summaries, and a submission manifest. Full
case-level public source URLs are stored in the benchmark case file. Hidden
outcomes and scoring notes are not included in the public ancillary case file,
so the benchmark can remain usable for future blinded evaluation. The project
should be described as a Codex CLI agent-harness benchmark unless rerun through
a raw model API harness.

\section*{Ethical and Safety Considerations}

The benchmark uses public historical sources and does not involve human
subjects, patient-level data, or wet-lab execution. The outputs should not be
used as medical, investment, regulatory, or clinical advice. The intended use
is evaluation of AI-agent organizational design. Real deployment would require
expert human oversight, audit trails, source verification, conflict-of-interest
controls, model monitoring, and explicit separation between research support
and regulated decision authority.

\end{document}